\pdfoutput=1
\documentclass{article}
\usepackage{preprint,times}
\usepackage{hyperref}
\usepackage{url}

\usepackage{multirow}
\usepackage{comment}

\usepackage[utf8]{inputenc}
\usepackage[T1]{fontenc}
\usepackage{graphicx}
\usepackage{pifont}
\newcommand{\cmark}{\ding{51}}
\newcommand{\xmark}{\ding{55}}
\usepackage{booktabs}
\usepackage{amsmath}
\usepackage{amsfonts}
\usepackage{nicefrac}
\usepackage{microtype}
\usepackage{xcolor}
\IfFileExists{axessibility.sty}{\usepackage[accsupp]{axessibility}}{}
\usepackage{xspace}
\usepackage{cleveref}
\usepackage{bm}
\usepackage{subfig}

\usepackage[normalem]{ulem}
\newcommand{\revcolor}{}
\newcommand{\old}[1]{}
\newcommand{\new}[1]{#1}
\newcommand{\rev}[1]{#1}
\newcommand{\revb}[1]{#1}
\newif\ifshowrevm
\showrevmfalse
\DeclareRobustCommand{\revm}[1]{\ifshowrevm\textcolor{blue}{#1}\else#1\fi}
\newcommand{\revmcolor}{\ifshowrevm\color{blue}\fi}

\newcommand{\ie}{\emph{i.e.}\xspace}

\newcommand{\vs}{\emph{vs.}\xspace}

\newcommand{\Oursname}{\textsc{AnchorComplete}\xspace}

\newcommand{\heading}[1]{\vspace{2pt}\noindent\textbf{#1.}}

\definecolor{bestC}{HTML}{B9D3EA}
\definecolor{secondC}{HTML}{D6E5F3}
\definecolor{thirdC}{HTML}{ECF3FA}

\title{\new{\textsc{AnchorComplete}}: Recovering Complete Scene Meshes from Partial Point Clouds}

\author{Weirong Chen$^{1,2*}$ \quad
Ismail El-Yamany$^{3*}$ \quad
Ganlin Zhang$^{1,2}$\\[0.2em]
\textbf{Chuanxia Zheng$^{3\#}$ \quad
Andrea Vedaldi$^{4}$ \quad
Daniel Cremers$^{1,2}$} \\[0.5em]
\normalfont 
$^{1}$Technical University of Munich \quad 
$^{2}$Munich Center for Machine Learning \quad \\
$^{3}$Nanyang Technological University \quad
$^{4}$University of Oxford \\[0.3em]
Project page: \href{https://wrchen530.github.io/anchor-complete}{\small\nolinkurl{wrchen530.github.io/anchor-complete}}
}

\begin{document}
\maketitle

\begingroup
\renewcommand{\thefootnote}{}
\footnotetext{$*$ Core technical contribution. \quad $\#$ Project lead.}
\endgroup

\vspace{-1cm}
\begin{figure}[htbp]
\centering
\includegraphics[width=\linewidth]{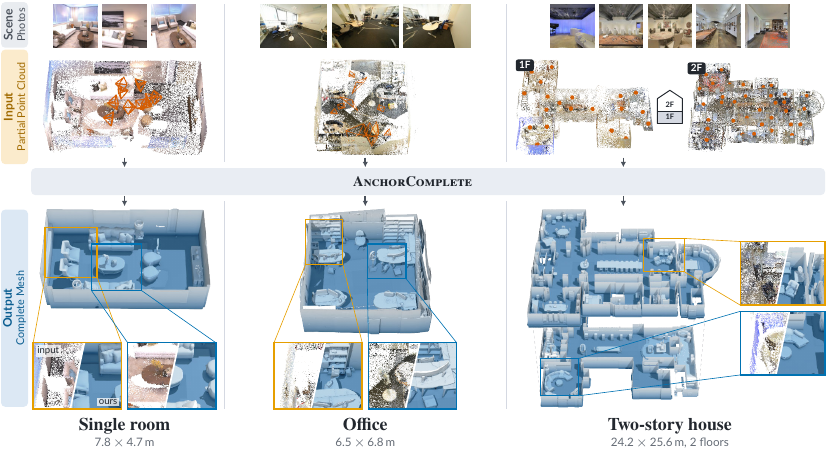}
\caption{\textbf{Completed scene meshes generated by \Oursname{}.}
 From a sparse, noisy point cloud, \Oursname{} recovers a complete, metric-scale mesh,
 shown with the house's floors pulled apart and insets that pair the input with our mesh.
 One model covers all three scales and fills in what the input never observed.
}
\label{fig:teaser}
\end{figure}

\begin{abstract}
Most image-based 3D reconstructors primarily recover visible surfaces,
producing partial point clouds with large holes and missing areas,
while leaving geometry outside the observed views unresolved. 
To address this gap, we introduce \Oursname{}, a framework that reconstructs complete scene meshes from partial point reconstructions.
Our design addresses two key challenges: 
anchoring generation to the observed geometry and scaling completion to \revm{scenes of different scales}.
First, 
we introduce a \emph{partial-point conditioning} module that combines global scene tokens with local spatial evidence to anchor generation to the input point cloud.
Second, we introduce \emph{metric-scale coarse-to-fine generation},
which establishes a global scene layout and progressively refines it within windows of fixed physical size.
Each window inherits the coarser geometry, preserving scene structure while recovering fine details at any scene size.
We instantiate these designs on top of
TRELLIS.2.
The resulting model produces complete scene meshes that remain faithful to the observed geometry, providing a more coherent digital representation of the physical environment.
Experiments on Replica and SCRREAM show that \Oursname{} consistently outperforms strong baselines in complete-scene reconstruction, particularly under sparse-view settings.

\end{abstract}

\section{Introduction}%
\label{sec:intro}

We study the problem of recovering a complete 3D mesh of a scene given only a partial observation, such as the output of a 3D reconstructor or a 3D scanner.
A robust solution to this problem would benefit downstream applications such as novel-view AR/VR rendering~\citep{xie26lavr}, large-scale physical simulation~\citep{habitat19iccv,szot21habitat}, and robotics~\citep{liu23libero:}, where complete scene meshes are often required for rendering, collision detection, and physical interaction.

Addressing this problem requires tackling two coupled challenges:
(i) preserving fidelity to the observed geometry and
(ii) scaling completion to any physical scene size.
Recently, feed-forward 3D reconstructors like DUSt3R~\citep{wang24dust3r:} and its variants~\citep{leroy24grounding,wang2025continuous,wang25vggt,yang25fast3r:,sun24sc-depthv3:,keetha25mapanything:,wang26vggt-omega}, as well as online SLAM systems like DROID-SLAM~\citep{teed21droid-slam:} and VGGT-SLAM~\citep{maggio25vggt-slam:}, have made it possible to obtain dense point clouds of scenes in a single forward pass by scaling the model and training data to directly regress 3D points from image observations.
However, completion is not part of their formulation, so unobserved regions remain large holes,
limiting their applicability in many downstream applications.

Early completion methods~\citep{han17high-resolution,lyu21a-conditional,yu21pointr:,zhou213d-shape} focus on completing point clouds of single objects in normalized coordinate spaces and thus do not scale to scene-level inputs.
While Amodal3R~\citep{wu25amodal3r}, SAM 3D~\citep{team25sam-3d:-3dfy}, ShapeR~\citep{siddiqui2026shaper},
and CoLa3D~\citep{chen2026cola3d} extend foundational 3D generators~\citep{zhang233dshape2vecset:,zhang24clay:,hunyuan3d25hunyuan3d,xiang24structured,xiang25native,li2025triposg} to occluded geometry, they mainly address separate \emph{object} completion in normalized coordinate spaces and do not directly condition on partial point observations, so they may not remain faithful to the geometry actually observed by the cameras.
Recent \emph{scene} completion methods like PaSCo~\citep{cao2024pasco}, NOVA3R~\citep{chen26nova3r}, and Seen2Scene~\citep{meng2026seen2scene} aim to generate complete scene geometry.
While promising, these models typically complete within the given field of view, without considering geometry outside the camera view.
They also do not output a complete mesh, but rather a point cloud or a voxel grid, which may not be directly usable for downstream applications.

Seeking to close this gap, we introduce \Oursname, a generative framework that takes \new{as input a partial point cloud, such as one recovered from sparse RGB observations,} and outputs a complete scene mesh \new{(\Cref{fig:teaser})}.
\new{We build on the TRELLIS~\citep{xiang24structured,xiang25native}, but adapting it to scene completion poses two key challenges: faithfully preserving observed geometry and scaling to \revm{scenes of varying spatial extent. When an entire scene is normalized into a single cube, the physical size of each mesh cell increases with the scene extent}.}

\new{We address these challenges with two main components. \emph{Partial-point conditioning} anchors generation to the input through a hybrid scheme that combines \new{global cross-attention to transformer-encoded anchor tokens with local spatial injection} of observed voxels into the noisy latent.
\revm{\emph{Metric-scale coarse-to-fine generation} covers the scene with an octree of fixed physical size windows, each refining its coarser parent, which keeps the layout consistent across windows while adding detail at any scene size.} Together, they enable faithful and scalable scene completion from sparse observations.}

\new{We evaluate our method on the Replica~\citep{straub19the-replica} and SCRREAM~\citep{jung2024scrreamscanregister} datasets, where we reconstruct complete indoor scenes under standard and sparse-view settings.
Our experiments demonstrate that \Oursname{} recovers more complete scene geometry than the baselines while remaining faithful to the visible input, and degrades more slowly as camera coverage drops. Our contributions are:}
\begin{itemize}
    \item We take a step towards recovering complete meshes of scenes from partial point clouds, without requiring densely sampled posed images, oriented normals, or per-scene optimization.
    For this new problem setting, we formulate strong baselines and evaluation protocols.
    \item We introduce \Oursname, a generative framework that contributes two architectural components: (1) \textbf{partial-point conditioning} and (2) \revm{\textbf{metric-scale coarse-to-fine generation} on an octree of fixed physical size windows, which decouples mesh resolution from scene extent}; together they enable generative mesh completion at scene scale while remaining anchored to the observed reconstruction.
    \item \Oursname{} achieves the best $\mathrm{CD}_{\mathrm{all}}$ and F-scores on Replica and SCRREAM, and degrades more slowly than the visible-surface baselines we evaluate as input views decrease.
\end{itemize}

\section{Related Work}%
\label{sec:related}

\paragraph{Visible-surface reconstruction.}
DUSt3R~\citep{wang24dust3r:} and MASt3R~\citep{leroy24grounding} regress dense pixel-aligned point maps from unposed image collections, marking a shift toward generalizable, pose-free 3D reconstruction.
Building on this paradigm, several recent works~\citep{wang25vggt,wang2025continuous,lin25depth, yang25fast3r:,sun24sc-depthv3:,keetha25mapanything:,wang2025pi, chen2026geometric,wang26vggt-omega} extend it to multi-view settings, while online dense SLAM systems~\citep{teed21droid-slam:,murai2025mast3r,liu2025slam3r,maggio25vggt-slam:} integrate such reconstruction into camera-trajectory estimation.
Surface fitters such as Poisson and Screened Poisson reconstruction~\citep{kazhdan2006poisson,kazhdan2013screened}, and learned methods such as NKSR and NoKSR~\citep{huang2023neural,li25noksr:}, convert oriented points into smooth implicit surfaces.
\new{All remain tied to camera-visible geometry: occluded regions go unmodeled, and surfaces fitted across holes follow smoothness rather than any prior over the unobserved scene. We instead complete those regions with a learned generative prior.}

\paragraph{Generative 3D priors.}
Generative 3D methods synthesize complete geometry from images, text, layouts, or latent representations. Image-space methods such as Text2Room~\citep{hollein23text2room:}, Invisible Stitch~\citep{engstler25invisible}, and GenRC~\citep{li24genrc:} iteratively inpaint and lift RGB-D views, while foundation 3D generators~\citep{zhang233dshape2vecset:,zhang24clay:,xiang24structured,hunyuan3d25hunyuan3d,li2025triposg,xiang25native} and amodal methods~\citep{wu25amodal3r,team25sam-3d:-3dfy,siddiqui2026shaper} recover complete object geometry from image or text prompts.
Scene-scale approaches~\citep{shriram25realmdreamer:,wu24blockfusion:,ju2024diffindscene,roessle24l3dg:,bahmani2023cc3d,yang24physcene:,yang24holodeck:,feng2023layoutgpt} generate rooms from learned scene representations or layouts.
However, these priors are not directly constrained by the metric geometry of a specific partial reconstruction, motivating our point-conditioned formulation.

\paragraph{Completion from partial 3D geometry.}
Completion methods condition on partial 3D input to infer missing geometry. Object-centric approaches~\citep{yuan2018pcn,yu21pointr:,zhou2022seedformer,chu2023diffcomplete,han17high-resolution,lyu21a-conditional,zhou213d-shape} complete point clouds or shapes, while scene methods~\citep{dai2018scancomplete,dai2020sgnn} predict occupancy or SDFs from partial voxel grids, typically at limited resolution.
More recent generative models such as Hunyuan3D-Omni~\citep{hunyuan3d2025hunyuan3d} and ShapeR~\citep{siddiqui2026shaper} condition foundation 3D generators on partial geometry but remain object-scale.
At scene scale, PaSCo~\citep{cao2024pasco}, NOVA3R~\citep{chen26nova3r}, Points-to-3D~\citep{xia2026points}, Seen2Scene~\citep{meng2026seen2scene}, GenRecon~\citep{schmid2026genrecon}
and Surflo~\citep{guedon2026surflo} explore occupancy, amodal reconstruction, latent generation, or chunk-based completion.
Seen2Scene is the closest baseline to our setting.
Maintaining fidelity to observed geometry while plausibly completing missing regions remains a central challenge in scene completion. We condition a pretrained 3D generative prior on a raw scene-scale partial reconstruction and output a completed mesh in the original scene coordinate frame (\Cref{sec:method}).

\section{Method}%
\label{sec:method}

\begin{figure}[tb!]
\centering
\includegraphics[width=\linewidth]{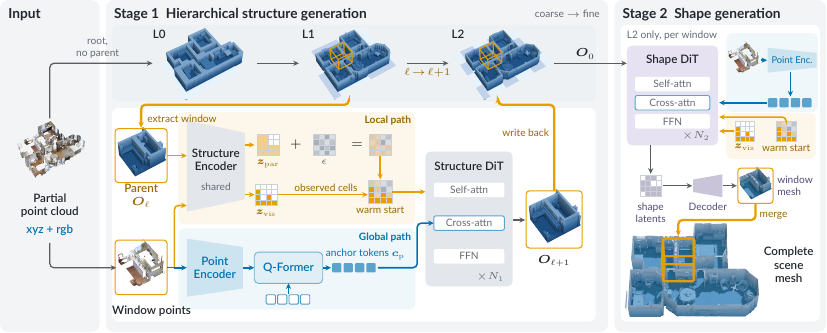}
\vspace{-10pt}

\caption{\textbf{Overview of \Oursname.} \revm{ From a partial, colored point cloud, Stage 1 generates a metric-scale octree coarse to fine, one window at a time, and Stage 2 generates shape latents on the finest level, which are decoded, merged and meshed once. Each window is conditioned on its parent occupancy and visible points through the local path (warm start) and on its points through anchor tokens read by cross-attention (global path).}
}
\label{fig:pipeline}
\vspace{-12pt}
\end{figure}

We develop \Oursname, a generative model that recovers a complete scene mesh $\mathcal{M}$ from a partial point cloud $\mathcal{P}$ obtained from feed-forward 3D reconstruction~\citep{wang25vggt,lin25depth}, SLAM~\citep{teed21droid-slam:,maggio25vggt-slam:}, or 3D scanning~\citep{dai17scannet:}.
The key challenge is to hallucinate plausible geometry in unobserved regions while preserving the observed structure.
Our model, illustrated in~\Cref{fig:pipeline}, builds on TRELLIS.2~\citep{xiang25native} and adapts it to scene-level completion through partial-point conditioning and \revm{metric-scale coarse-to-fine generation}.
We first review TRELLIS in \Cref{sec:prelim}, then present \revm{partial-point conditioning in \Cref{sec:vpc} and the coarse-to-fine octree that extends the generator from a single normalized cube to metric scale in \Cref{sec:metric},} followed by training details in \Cref{sec:training}.

\subsection{Preliminaries: TRELLIS}%
\label{sec:prelim}

TRELLIS~\citep{xiang24structured} is a foundational 3D asset generator that takes a text prompt or image as the condition $\mathbf{c}$ and outputs various 3D representations,
including meshes $\mathcal{M}$,
Gaussian splats $\mathcal{G}$,
and radiance fields $\mathcal{R}$.
For flexibility, TRELLIS represents the given 3D asset $\mathcal{O}$ with a set of \emph{Structured Latents} (SLATs) $\mathcal{Z} = \{\bm{z}_i, \bm{p}_i\}_{i=1}^{L}$ on a predefined $H\times W\times D$ spatial grid $\Omega$, where $L\ll HWD$ is the number of active voxels, $\bm{p}_i \in \mathbb{R}^3$ is the 3D coordinate of the $i$-th active voxel, and $\bm{z}_i \in \mathbb{R}^C$ is the corresponding latent feature encoding local geometry and appearance information.

The TRELLIS pipeline consists of two VAEs and two corresponding rectified-flow generation stages~\citep{albergo2022building,lipman22flow,liu23flow}.
The \emph{structure} VAE $(\mathcal{E}_\text{s}, \mathcal{D}_\text{s})$ compresses a dense binary voxel grid $\bm{O}\in\{0,1\}^{H\times W\times D}$ into a latent representation $\bm{z}_\text{s}\in\mathbb{R}^{\frac{H}{s}\times\frac{W}{s}\times\frac{D}{s}\times C}$, which is decoded to model the coarse geometry.
The \emph{SLAT} VAE $(\mathcal{E}_\text{sl}, \mathcal{D}_\text{sl})$ encodes the \emph{Structured Latents} (SLATs) $\mathcal{Z}$ together with voxelized multi-view image features into $\bm{z}_\text{sl}\in\mathbb{R}^{L\times C}$, which is decoded into various 3D representations.

During generation, a flow transformer $\mathcal{T}_\text{s}$ first generates the structure latent $\bm{z}_\text{s}$ from Gaussian noise conditioned on text or image tokens $\mathbf{c}$.
The sampled latent is decoded into the active voxel support $\{\bm{p}_i\}_{i=1}^{L}$, which determines the coarse spatial layout.
A sparse flow transformer $\mathcal{T}_\text{sl}$ then generates the local SLAT features $\{\bm{z}_i\}_{i=1}^{L}$ on this support, again conditioned on $\mathbf{c}$, and the resulting SLAT $\mathcal{Z}$ is decoded into the desired 3D output.
TRELLIS.2~\citep{xiang25native} retains this two-stage design but replaces the second-stage SLAT features with an O-Voxel shape latent.
We build on TRELLIS.2 and extend it with partial-point conditioning and scalable scene-level generation.

\subsection{Partial Point Conditioning}%
\label{sec:vpc}

Unlike image-conditioned 3D generators~\citep{zhang24clay:,hunyuan3d25hunyuan3d,xiang24structured,xiang25native},
we condition scene completion directly on partial point observations.
This is advantageous for several reasons:
(1) \emph{Completeness}: the observed points delineate the visible scene boundary, providing direct geometric evidence of where completion is needed;
(2) \emph{Accuracy}: the condition is directly grounded in 3D geometry, maintaining better fidelity to the visible geometry;
(3) \emph{Efficiency}: the observed points $\mathcal{P}=\{\bm{p}_i\}_{i=1}^{N}$ can be uniformly subsampled to a fixed size $N$, avoiding conditioning costs that grow with the number of input images;
and (4) \emph{Composability}: point-based inputs can be chunked for scalable local processing, and the outputs can be directly merged in the same scene coordinate frame.
In particular, we take a hybrid conditioning approach via (1) \new{global cross-attention to \emph{anchor tokens} encoded from the point cloud} and (2) \new{\emph{local spatial injection} of visible-voxel evidence into the structure latent}\new{;
both paths are shown in \Cref{fig:pipeline}}.
The two pathways are ablated in \Cref{sec:experiment}.

\heading{\new{Global Anchor Tokens}}
We design a simple architecture to extract \new{anchor tokens} from the input point observations $\mathcal{P}=\{\bm{p}_i\}_{i=1}^{N}$.
In particular, we implement the point-cloud encoder \rev{$\mathcal{E}_\text{p}$} using a transformer architecture inspired by the point encoder of NOVA3R~\citep{chen26nova3r}, but with its per-point features taken from a pretrained Utonia backbone~\citep{zhang26utonia:}.
To this end, we first extract high-dimensional features $\{\bm{f}_i\}_{i=1}^{N}$ for all $N$ input points with the frozen Utonia backbone.
The encoder then uses $m$ learnable queries $\mathcal{F}^q$ and the full set of point features $\mathcal{F}^{(k,v)}=\{\bm{f}_i\}_{i=1}^{N}$ as the keys and values for the encoder's cross-attention layers, where $m \ll N$ is a fixed number of point tokens.
The \new{anchor tokens} are further refined by several interleaved self-attention and cross-attention blocks.
The resulting \new{anchor tokens} $\bm{c}_\text{p}\in \mathbb{R}^{m \times C_\text{xa}}$, where $C_\text{xa}$ is the cross-attention feature dimension, are then fed to the TRELLIS.2 flow model as global conditioning tokens through cross-attention, replacing the original image/text conditioning tokens in both stages.
\revm{We use $m=768$ anchor tokens. The encoder also reads the point colours and, in the metric-scale model of \Cref{sec:metric}, two per-point attributes: an observed-point flag and the expected depth-noise level.}

\heading{\new{Local Spatial Injection}}
To further encourage spatial consistency between the generated geometry and the input point cloud, we introduce a local spatial conditioning path that injects visible-point evidence directly into the first-stage structure latent space.
Inspired by condition injection strategies in image editing~\citep{avrahami2023blended}, we first voxelize the point observations $\mathcal{P}$ into a binary 3D grid $\bm{O}_{\mathrm{vis}} \in \{0,1\}^{H \times W \times D}$, setting a voxel to 1 if it contains at least one observed point and to 0 otherwise.
It is then encoded with the pretrained structure encoder $\mathcal{E}_\text{s}$ to obtain partial-geometry features $\bm{z}_{\mathrm{vis}} \in\mathbb{R}^{\frac{H}{s}\times \frac{W}{s}\times \frac{D}{s}\times C}$.
In parallel, we downsample $\bm{O}_{\mathrm{vis}}$ to obtain a 3D binary mask $\bm{O}_{\mathrm{vis},m} \in \{0,1\}^{\frac{H}{s} \times \frac{W}{s} \times \frac{D}{s}}$, indicating which latent cells correspond to visible regions.
The visible evidence enters Stage 1 in two ways.
First, it initializes the flow: rather than starting from Gaussian noise everywhere, we replace the initial sample at visible cells by $\bm{z}_{\mathrm{vis}}$,
\begin{equation}
  \label{eq:warmstart}
  \bm{z}_s^{(1)} \;=\; \bm{O}_{\mathrm{vis},m} \odot \bm{z}_{\mathrm{vis}} \;+\; (\mathbf{1} - \bm{O}_{\mathrm{vis},m}) \odot \bm{\epsilon},
  \qquad \bm{\epsilon} \sim \mathcal{N}(\mathbf{0}, \bm{I}),
\end{equation}
where $\bm{z}_s^{(1)}$ is the latent the sampler starts from.
Second, the mask is available at every denoising step: we concatenate $\bm{O}_{\mathrm{vis},m}$ to the noisy latent and feed the result back to the denoiser, so it can separate observed from unobserved cells throughout sampling.
\revm{Stage 2 uses both as well: it receives $\bm{z}_{\mathrm{vis}}$, read at its voxels, with a per-voxel visibility flag at every denoising step, and its initial sample takes, on observed voxels, the shape latents encoded from the visible points.}

{\revmcolor
\subsection{Metric-Scale Coarse-to-Fine Generation}%
\label{sec:metric}

A single normalized cube is stretched to fit the whole input, so its cells grow with the scene. In a small room this hardly matters, but in a large apartment each cell ends up several centimetres wide and thin parts such as chair legs or shelf boards disappear. A finer cube does not help either, since a resolution that suits the apartment is wasted on a bathroom.
We therefore stop tying the cell size to the scene and generate it coarse to fine.

\begin{figure}[t]
\centering
\includegraphics[width=\linewidth]{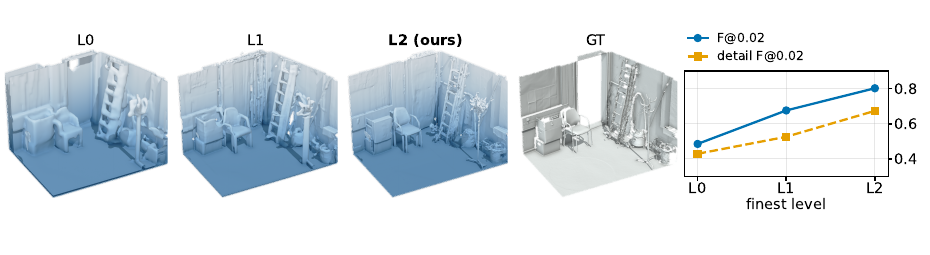}
\vspace{-14pt}
\caption{\revm{\textbf{Coarse-to-fine levels.} The same model stopped at levels 0, 1 and 2 on a room corner of a real-world scene, with the mean F-scores of \Cref{tab:ablation}(b) on the right. Only level 2 separates the ladder rungs and the legs of the chair.}}
\label{fig:octree_depth}
\vspace{-8pt}
\end{figure}

\heading{Multi-Levels}
Level~0 is the single cube itself: it sees the whole scene at once and gives the overall layout of the rooms, even if the details are rough.
Levels~1 and~2 then cover the occupied space with windows of a fixed physical size, the level-2 windows half as wide as the level-1 windows, so each level's $512^3$ mesh grid has a fixed cell size regardless of scene size, and further levels $3,4,\dots$ would halve the window again; \Cref{fig:octree_depth} shows what each level resolves. In observed regions, the window follows the input geometry; in unobserved regions, it is initialized from the parent prediction and primarily refines that prediction rather than synthesizing a new layout.
Inside a window, both stages of \Cref{sec:vpc} run unchanged on the window's crop of the input; the model never needs to know how large the scene is. All levels use the sparse-structure and O-Voxel latents of TRELLIS.2~\citep{xiang25native}.

\heading{Parent warm start}
A window at level $\ell{+}1$ does not start from scratch. We encode its parent's occupancy, $\bm{z}_{\mathrm{par}}=\mathcal{E}_\text{s}(\bm{O}_{\ell})$, and use it as the starting point wherever the input has no points:
\begin{equation}
  \label{eq:parentwarm}
  \bm{z}_s^{(1)} \;=\; \bm{O}_{\mathrm{vis},m} \odot \bm{z}_{\mathrm{vis}} \;+\; (\mathbf{1} - \bm{O}_{\mathrm{vis},m}) \odot \big(\alpha\, \bm{z}_{\mathrm{par}} + \sqrt{1-\alpha^2}\, \bm{\epsilon}\big),
  \qquad \bm{\epsilon} \sim \mathcal{N}(\mathbf{0}, \bm{I}),
\end{equation}
with $\alpha=0.8$ ($\alpha=0$ recovers \Cref{eq:warmstart}).
Where the input saw the scene, the window follows the observation; where it did not, the window starts from the parent's guess and mostly sharpens it instead of inventing a new room. Since every window inherits the same coarse layout, neighbouring windows agree on where the walls and furniture are.
Windows overlap by a quarter of their width and keep only their centre. At the finest level we run Stage~2, merge the decoded windows into a common grid and extract a single mesh, thereby avoiding explicit mesh stitching.
}

\subsection{Training}%
\label{sec:training}

We fine-tune two flow models in TRELLIS.2 with our new conditioning: one predicts the active voxels $\{\bm{p}_i\}_{i=1}^{L}$, and the other predicts the shape latents $\{\bm{z}_i\}_{i=1}^{L}$ on this support.
Each model $v_{\bm{\theta}}$ operates in a latent space $\bm{z}$ and is trained to remove Gaussian noise $\boldsymbol{\epsilon} \sim \mathcal{N}(\mathbf{0},\mathbf{I})$ added to the target latent $\bm{z}_0$, \ie $\bm{z}_t = (1-t)\bm{z}_0 + t\boldsymbol{\epsilon}$, where $t \sim \mathcal{U}(0,1)$ is a random interpolation timestep.
Formally, each model $v_{\bm{\theta}}$ is trained to predict the rectified-flow velocity:
\begin{equation}
  \label{eq:loss}
  \mathcal{L}(\bm{\theta})
  =
  \mathbb{E}_{(\bm{z}_0, \mathbf{c}_p), t,\boldsymbol{\epsilon}}
  \left[
  \left\|
  v_{\bm{\theta}}(\bm{z}_t, \mathbf{c}_p, t)
  -
  \left(\boldsymbol{\epsilon} - \bm{z}_0\right)
  \right\|_2^2
  \right],
\end{equation}
where $\mathbf{c}_p$ denotes the encoded partial point observations.
The loss is applied independently to both stages, and we fine-tune the newly introduced conditioning modules \revm{and the denoisers} from the pretrained generative prior.
The warm start of \Cref{eq:warmstart} and the mask concatenation are active during both training and inference.
\revm{For the metric-scale model (\Cref{sec:metric}), both stages are trained on windows: each sample crops a window and the depth points of 2--8 views that see it, drawing levels~1 and~2 most often. The parent of \Cref{eq:parentwarm} is the next coarser level: usually ($p=0.7$) a prediction (the single cube for level~1, an earlier model's for level~2), otherwise a damaged ground-truth parent, so the model learns to correct imperfect parents rather than copy them.}

\section{Experiments}%
\label{sec:experiment}

\subsection{Experimental Setup}%
\label{sec:exp:setup}

\heading{Evaluation datasets}
\new{We evaluate on two indoor datasets with ground-truth geometry for both visible and occluded regions: \textbf{Replica}~\citep{straub19the-replica} with 8 photorealistic indoor reconstructions with high-quality meshes and densely sampled trajectories, and \textbf{SCRREAM}~\citep{jung2024scrreamscanregister} with 12 real-world captures with accurate per-scene ground truth, which we treat as our primary real-world benchmark.
Note that, we do not use any of the training scenes from these datasets.
}

\heading{Training data construction}
\new{We train \Oursname{} on \textbf{InternScenes}~\citep{zhong2025internscenes} and \textbf{SAGE}~\citep{Xia2026sage}. We use around $16\text{k}$ scenes from InternScenes, whose watertight glTF meshes serve as complete supervision $\mathcal{M}$, and around $7\text{k}$ rooms from SAGE, whose complete meshes are used the same way; both cover observed surfaces and occluded regions. Per scene we render multi-view RGB-D (InternScenes with BlenderProc~\citep{denninger23blenderproc2:}) at $640{\times}480$ and an $85^\circ$ horizontal field of view, placing cameras with an area-adaptive multi-anchor sampler plus corner viewpoints for coverage. Details are in the appendix.}

\heading{Metrics}
\new{Following prior work~\citep{wang24dust3r:,wang25vggt,chen26nova3r,guedon2026surflo}, we report Chamfer Distance (CD) and F-score, using ground-truth poses to crop the \textbf{visible} mesh within the input view frustums. We decompose CD into \emph{accuracy} (prediction-to-GT) and \emph{completeness} (GT-to-prediction). $\mathrm{CD}_{\mathrm{vis}}$ is symmetric over the visible region, while $\mathrm{CD}_{\mathrm{all}}$ combines visible-region accuracy with full-scene completeness. F-scores at $0.1$ and $0.05\,\mathrm{m}$ use the same regions as $\mathrm{CD}_{\mathrm{all}}$; lower CD and higher F are better.}
Accuracy is computed only for predictions inside the input-view frustums, since completion outside the observed region is inherently multi-modal and the ground truth represents only one valid realization.
See \Cref{sec:supp:impl} for details.

\begin{table*}[t]
\centering
\setlength{\tabcolsep}{4pt}
\footnotesize
\caption{\textbf{Quantitative reconstruction evaluation on Replica~\citep{straub19the-replica} and SCRREAM~\citep{jung2024scrreamscanregister} with $N_{\mathrm{views}} = 4$.}
The best result in each \new{metric} column is shown in \textbf{bold}, and the second best is \underline{underlined}.
\new{$^*$ denotes our scene finetune of TRELLIS, with its output aligned to the input cloud; the unstarred row is the released object-pretrained model.}
\new{Runtime is per-scene reconstruction on an RTX PRO 6000 Max-Q.
}}%
\label{tab:e1}
\resizebox{\linewidth}{!}{
\begin{tabular}{@{}l cccc cccc c@{}}
\toprule
\multirow{2}{*}{\textbf{Method}}                       & \multicolumn{4}{c}{\textbf{Replica}}
                       & \multicolumn{4}{c}{\textbf{SCRREAM}}
                       & \new{\multirow{2}{*}{\textbf{Runtime} $\downarrow$}} \\
\cmidrule(lr){2-5} \cmidrule(lr){6-9}
                 & $\mathrm{CD}_{\mathrm{vis}}$ $\downarrow$ & $\mathrm{CD}_{\mathrm{all}}$ $\downarrow$ & F$_{0.1}$ $\uparrow$ & F$_{0.05}$ $\uparrow$
                       & $\mathrm{CD}_{\mathrm{vis}}$ $\downarrow$ & $\mathrm{CD}_{\mathrm{all}}$ $\downarrow$ & F$_{0.1}$ $\uparrow$ & F$_{0.05}$ $\uparrow$ \\
\midrule
\multicolumn{10}{l}{\underline{\emph{Feed-forward dense reconstruction}}} \\
\revm{VGGT-$\Omega$}~\citep{wang26vggt-omega}  & \revm{0.062} & \revm{0.230} & \revm{0.560} & \revm{0.412} & \revm{0.062} & \revm{0.247} & \revm{0.582} & \revm{0.433} & \new{<1s} \\
DA3~\citep{lin25depth}                    & \textbf{0.048} & 0.220 & 0.611 & 0.508 & \textbf{0.052} & 0.227 & 0.607 & 0.453 & \new{<1s} \\
CUT3R~\citep{wang2025continuous}                  & 0.244 & 0.360 & 0.355 & 0.214 & 0.189 & 0.359 & 0.455 & 0.311 & \new{<1s} \\
PI-3~\citep{wang2025pi}                   & \underline{0.053} & 0.224 & 0.573 & 0.426 & 0.096 & 0.253 & 0.485 & 0.315 & \new{<1s} \\
\midrule
\multicolumn{10}{l}{\underline{\emph{Image-based Generation}}} \\
\new{TRELLIS}~\citep{xiang24structured}       & \new{0.537} & \new{0.584} & \new{0.239} & \new{0.127} & \new{0.346} & \new{0.416} & \new{0.262} & \new{0.160} & \new{$\sim$4 s} \\
\new{TRELLIS}$^*$~\citep{xiang24structured}   & \new{0.244} & \new{0.288} & \new{0.308} & \new{0.150} & \new{0.221} & \new{0.264} & \new{0.323} & \new{0.170} & \new{$\sim$4 s} \\
GenRC~\citep{li24genrc:}                  & 0.355     & 0.366    & 0.161     & 0.076     & 0.295    & 0.332     & 0.191    & 0.091 & \new{$\sim$6.6 min} \\
\revm{Surflo}~\citep{guedon2026surflo}    & \revm{0.162} & \revm{0.318} & \revm{0.442} & \revm{0.288} & \revm{0.094} & \revm{0.270} & \revm{0.482} & \revm{0.325} & \revm{$\sim$1.2 min} \\
\midrule
\multicolumn{10}{l}{\underline{\emph{Scene completion}}} \\
SG-NN~\citep{dai2020sgnn}                  & 0.085 & 0.214 & 0.600 & 0.415 & 0.111 & 0.225 & 0.518 & 0.308 & \new{18 s} \\
NKSR~\citep{huang2023neural}                   & 0.057 & \new{0.211} & \new{0.633} & \new{0.527} & 0.066 & \new{0.209} & \new{0.614} & \new{0.457} & \new{1 s} \\
\new{Seen2Scene}~\citep{meng2026seen2scene} & \new{0.070} & \new{\underline{0.132}} & \new{\underline{0.777}} & \new{\underline{0.647}} & \new{0.075} & \new{\underline{0.167}} & \new{\underline{0.692}} & \new{\underline{0.525}} & \new{$\sim$10.4 min} \\
\midrule
\textbf{\Oursname{}} (ours)
                       & \revm{0.055} & \revm{\textbf{0.067}} & \revm{\textbf{0.819}} & \revm{\textbf{0.684}} & \revm{\underline{0.057}} & \revm{\textbf{0.076}} & \revm{\textbf{0.786}} & \revm{\textbf{0.584}} & \revm{$\sim$3 min} \\
\bottomrule
\end{tabular}%
}
\vspace{-10pt}
\end{table*}

\begin{figure}[t]
\centering
\includegraphics[width=\linewidth]{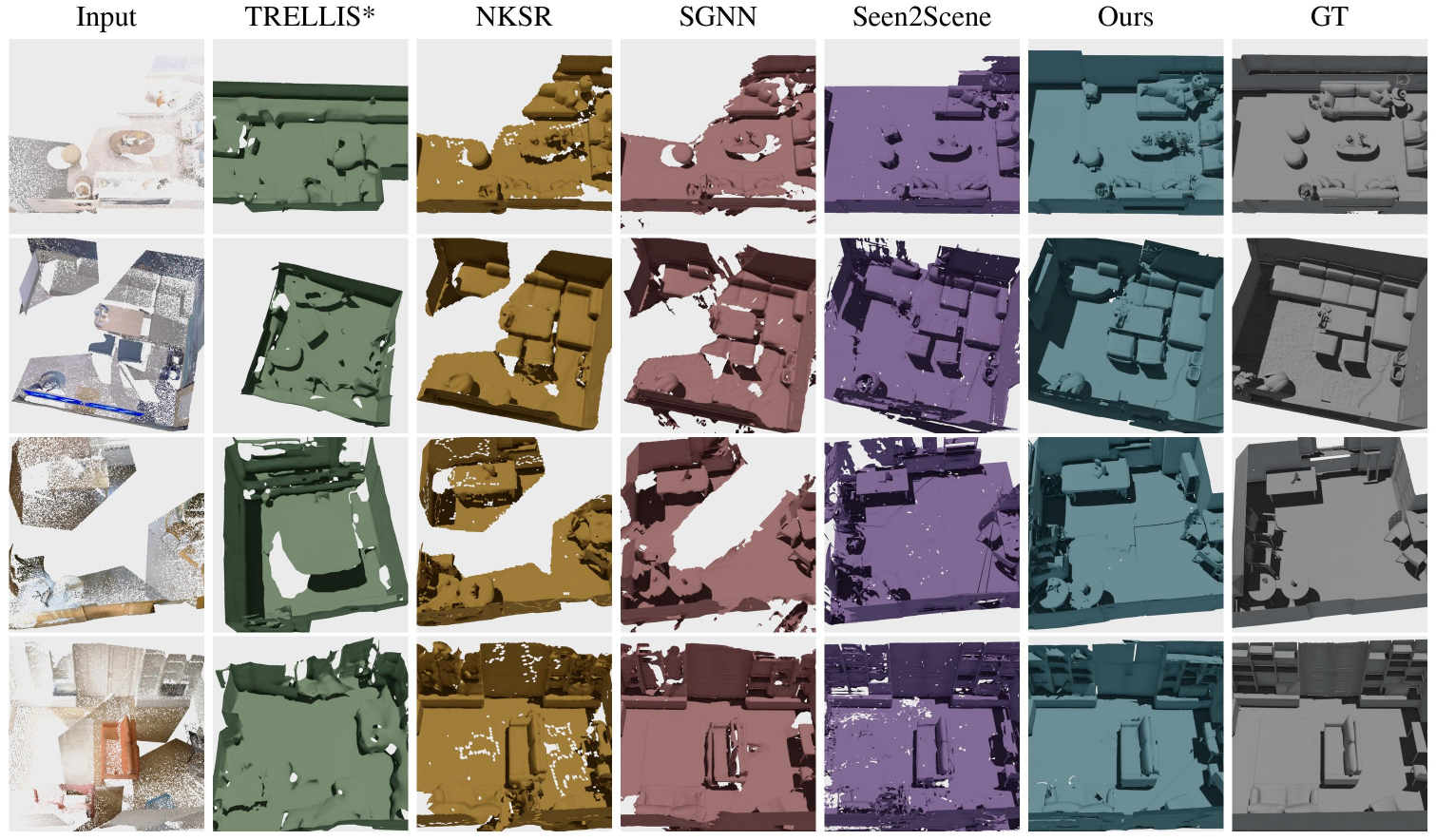}
\vspace{-10pt}
\caption{
\textbf{Qualitative reconstruction results on Replica~\citep{straub19the-replica} and SCRREAM~\citep{jung2024scrreamscanregister}.}
Given the same DA3 point cloud as input, our method achieves better scene completion performance.
}%
\label{fig:vis_qualitative}
\vspace{-10pt}
\end{figure}

\heading{Baselines}
\new{We compare against three groups, grouped as in \Cref{tab:e1}.
\emph{\textbf{i)} Feed-forward dense reconstruction}: VGGT-$\Omega$~\citep{wang26vggt-omega}, DA3~\citep{lin25depth}, CUT3R~\citep{wang2025continuous}, and PI-3~\citep{wang2025pi} predict dense geometry from images but target visible surfaces only; none outputs a mesh, so we convert all of them with the same pipeline.
\emph{\textbf{ii)} Image-based generation}: TRELLIS~\citep{xiang24structured}, GenRC~\citep{li24genrc:} and Surflo~\citep{guedon2026surflo} test whether generative priors alone recover missing geometry.
\emph{\textbf{iii)} Scene completion}: SG-NN~\citep{dai2020sgnn}, NKSR~\citep{huang2023neural}, and Seen2Scene~\citep{meng2026seen2scene} take partial point clouds and predict complete surfaces, matching our input-output setting most closely. We run Seen2Scene using the authors' released code and checkpoints. Its input is a TSDF volume fused from the same DA3 depth maps used by our model, using VDBFusion with $1.1\,\mathrm{cm}$ voxels. Both methods are evaluated using the same code after the same Sim(3) alignment.}

\heading{Implementation details}
We initialize both stages of \Oursname{} from the TRELLIS.2~\citep{xiang25native} image-to-3D checkpoints. The point-cloud encoder $\mathcal{E}_\text{p}$ keeps the Utonia backbone~\citep{zhang26utonia:} frozen; only the four-layer Q-Former that turns its features into the $m{=}768$ anchor tokens is trained. Both denoisers are 30-block DiTs with 1536 channels, 12 attention heads and rotary position embeddings. Stage~1 operates on the 8-channel $16^3$ sparse-structure latent, which is decoded into a $64^3$ occupancy grid and max-pooled to the $32^3$ support of Stage~2; Stage~2 generates the 32-channel O-Voxel latent on that support, from which the mesh is extracted on a $512^3$ grid.
Both stages are trained independently using AdamW with learning rate $2{\times}10^{-5}$, weight decay $0.01$, batch size 1 per GPU and $10\%$ condition dropout, each fine-tuned for 40k steps on four NVIDIA L40 GPUs.
At inference each stage takes 12 flow steps. Additional details are provided in the appendix.

\subsection{Complete Scene Reconstruction}%
\label{sec:exp:e1}

\Cref{tab:e1} evaluates complete-scene reconstruction at $N_{\mathrm{views}}{=}4$. \Oursname{} leads the complete-region metrics on both datasets against every baseline. On the visible region DA3~\citep{lin25depth} obtains the best $\mathrm{CD}_{\mathrm{vis}}$ with \Oursname{} close behind, so the completion gains do not cost visible-surface fidelity; CUT3R~\citep{wang2025continuous} struggles even there, likely because four views give little coverage of a full room.
Surflo~\citep{guedon2026surflo} can generate complete meshes, but it still lags behind \Oursname{} by a wide margin, likely because it is trained mainly on object-centric data.
Surface fitting such as NKSR~\citep{huang2023neural} smooths visible points into surfaces but still lags, showing smoothing alone cannot recover unseen geometry, while TRELLIS~\citep{xiang24structured} without partial-point conditioning drifts off the captured scene.
\rev{TRELLIS$^*$, our scene-level fine-tuned version of the same prior, improves substantially over the released object model but still trails \Oursname{} by a wide margin on both datasets, indicating that our advantage comes from conditioning on the observed geometry rather than from scene-level training.}

\Cref{fig:vis_qualitative} provides qualitative comparisons under the same four-view setting. \Oursname{} keeps the visible surface and closes back walls and out-of-view floor where the feed-forward and surface-fitting baselines leave holes. TRELLIS$^*$ returns a plausible room unrelated to the input cloud. \rev{Seen2Scene~\citep{meng2026seen2scene}, the completion baseline closest to our setting,} \revb{returns a far flatter result, without the furniture relief and wall structure \Oursname{} preserves.}

{\revmcolor
\begin{figure}[t]
\centering
\includegraphics[width=\linewidth]{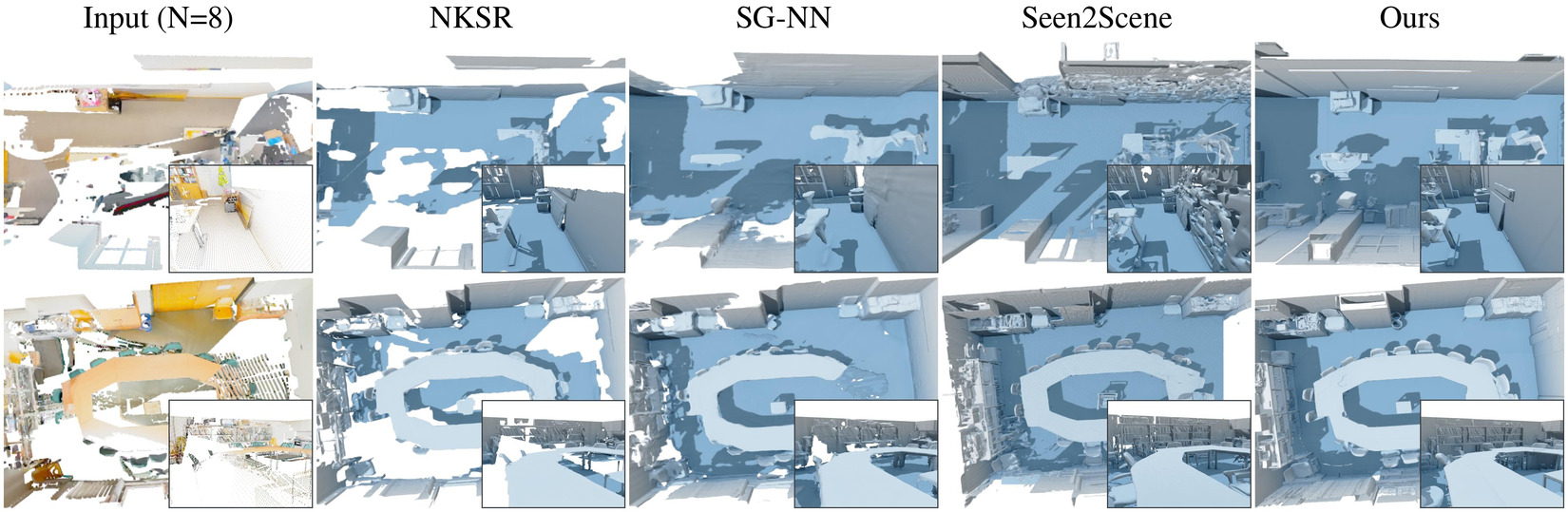}
\caption{\textbf{Real-world scenes.} Two real-world rooms from eight views, top-down with an inside view inset (same camera in every column); NKSR~\citep{huang2023neural}, SG-NN~\citep{dai2020sgnn}, Seen2Scene~\citep{meng2026seen2scene} and \Oursname{} take the same input. The baselines leave surfaces holed where coverage thins, while ours closes them.}%
\label{fig:realworld}
\end{figure}
}

\revm{\heading{Real-world scenes} \Cref{fig:realworld} takes the model outside the benchmarks, onto two real-world rooms reconstructed from eight views, with NKSR, SG-NN and Seen2Scene run on the same input. All three recover the room layout; the inside views are where they differ, as the baselines leave surfaces ragged and holed where coverage thins while ours closes them. These rooms have no complete ground-truth mesh, so they are shown qualitatively rather than scored.}

\begin{figure}[t]
\centering
\includegraphics[width=\linewidth]{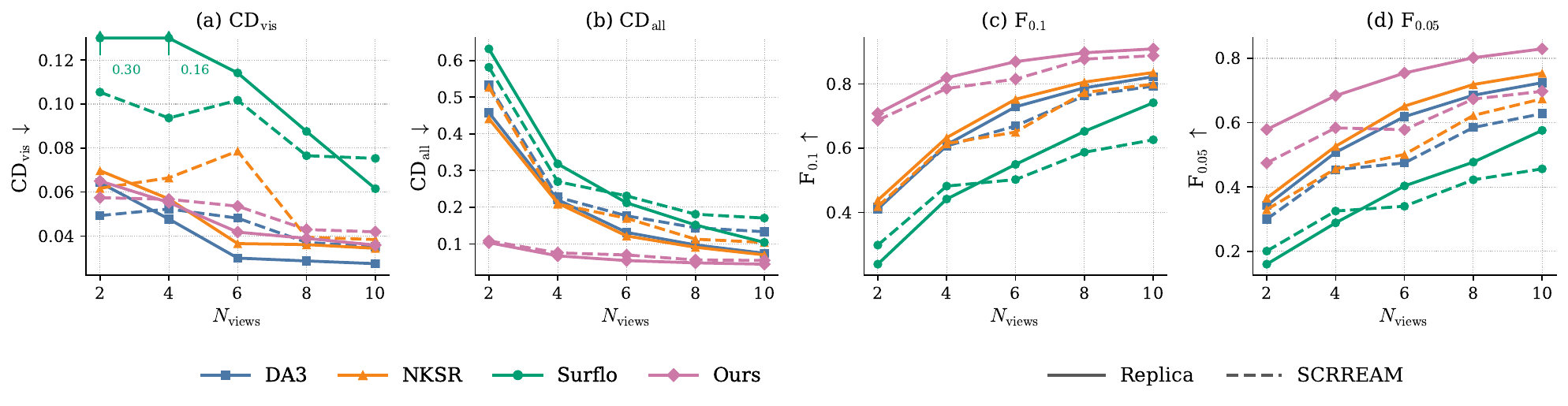}
\vspace{-8pt}
\caption{\textbf{Sparse-view robustness on Replica~\citep{straub19the-replica} and SCRREAM~\citep{jung2024scrreamscanregister}} (mean across scenes).}
\label{tab:e2}
\vspace{-9pt}
\end{figure}

\begin{figure}[t]
\centering
\includegraphics[width=\linewidth,trim=0 4mm 0 0,clip]{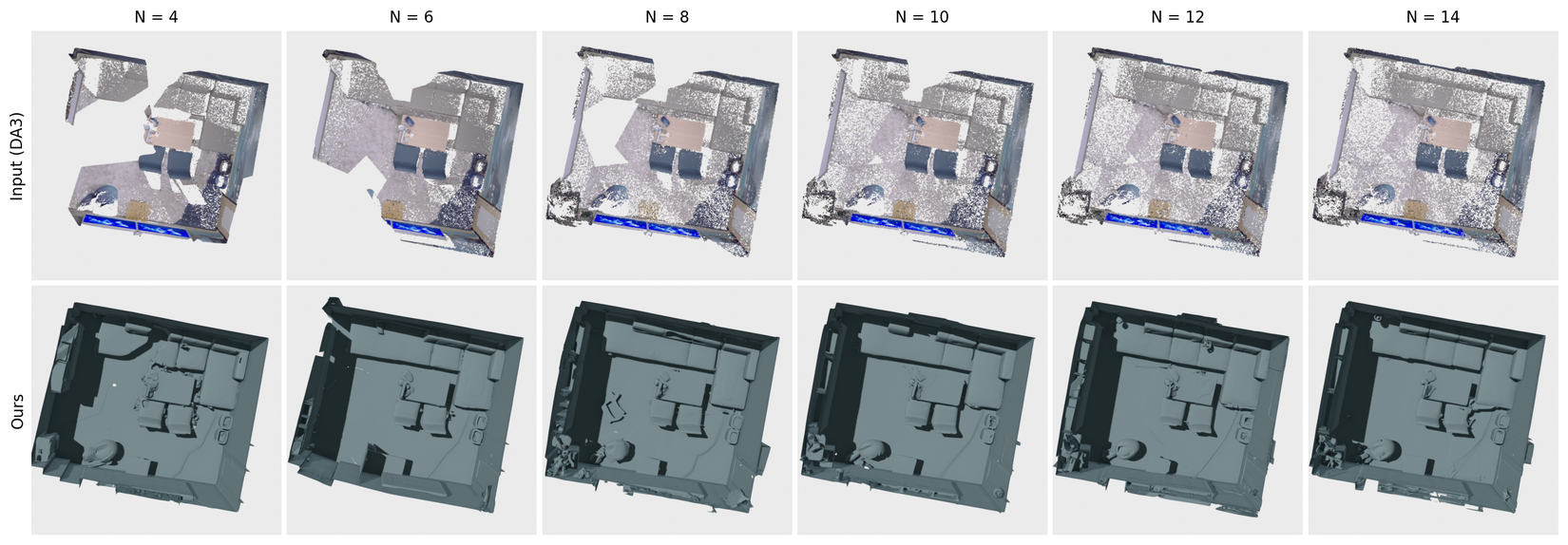}
\vspace{-8pt}
\caption{\textbf{Sparse-view conditional scene completion.} The DA3 input (top) loses coverage as views are removed, while our completion (bottom) stays closed across the range. \revm{Bottom row: \Oursname{} at level 0, $N{=}4$ to $14$ views; Replica provides 10 input frames per scene, so $N{=}12$ and $14$ add held-out frames of the same sequence.}}%
\label{fig:vis_sparse}
\vspace{-12pt}
\end{figure}

\begin{table}[t]
\centering
\footnotesize
\setlength{\tabcolsep}{3pt}
\renewcommand{\arraystretch}{0.95}
\caption{\textbf{Ablation study.} \revm{(a) Partial-point conditioning on Replica~\citep{straub19the-replica} ($N_{\mathrm{views}}{=}4$; the two ablated rows are trained on the TRELLIS backbone) and (b) the finest level of the coarse-to-fine generation on held-out scenes; distances in metres.}}
\label{tab:ablation}
\begin{minipage}[t]{0.60\linewidth}\vspace{0pt}\centering
\resizebox{\linewidth}{!}{\scriptsize\setlength{\tabcolsep}{6pt}%
\begin{tabular}{@{}l cc ccccc@{}}
\toprule
\multicolumn{8}{@{}l}{\revm{(a) \textbf{Partial-point conditioning}}} \\
\textbf{Method} & \textbf{Global} & \textbf{Local} & $\mathrm{CD}_{\mathrm{vis}}$ & $\mathrm{CD}_{\mathrm{all}}$ & F$_{0.1}$ & F$_{0.05}$ & F$_{0.02}$ \\
\midrule
TRELLIS$^*$ & \xmark & \xmark & 0.244 & 0.288 & 0.308 & 0.150 & 0.030 \\
\midrule
w/o local injection & \cmark & \xmark & 0.072 & 0.105 & 0.749 & 0.532 & 0.131 \\
w/o global tokens & \xmark & \cmark & 0.060 & 0.086 & 0.787 & 0.652 & 0.219 \\
\midrule
\revm{\textbf{Ours}} & \revm{\cmark} & \revm{\cmark} & \revm{\textbf{0.055}} & \revm{\textbf{0.067}} & \revm{\textbf{0.819}} & \revm{\textbf{0.684}} & \revm{\textbf{0.252}} \\
\bottomrule
\end{tabular}}
\end{minipage}\hfill
\begin{minipage}[t]{0.355\linewidth}\vspace{0pt}\centering\revmcolor
\resizebox{\linewidth}{!}{\scriptsize\setlength{\tabcolsep}{2pt}%
\begin{tabular}{@{}l ccccc@{}}
\toprule
\multicolumn{6}{@{}l}{(b) \textbf{Coarse-to-fine levels}} \\
\textbf{Finest} & Acc & Comp & F$_{0.05}$ & F$_{0.02}$ & F$^{\mathrm{det}}_{0.02}$ \\
\midrule
$\ell=0$ & 0.045 & 0.046 & 0.835 & 0.485 & 0.429 \\
$\ell=1$ & 0.038 & \textbf{0.036} & \textbf{0.873} & 0.675 & 0.526 \\
$\ell=2$ & \textbf{0.028} & 0.037 & 0.872 & \textbf{0.801} & \textbf{0.673} \\
\bottomrule
\end{tabular}}
\end{minipage}

\vspace{-12pt}
\end{table}

\begin{figure}[t]
\centering
\includegraphics[width=\linewidth]{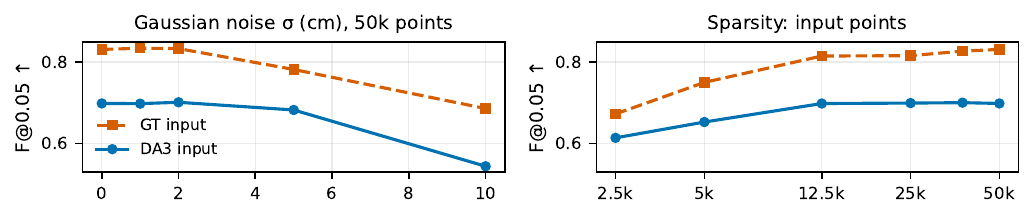}
\vspace{-6pt}
\caption{\textbf{Input robustness.} F$_{0.05}$ as the input is perturbed with Gaussian noise at a fixed 50k points
(left) or thinned out (right), with no retraining, each on the DA3 cloud and on ground-truth points. The model is
unaffected by noise up to the voxel scale and by sparsity down to a quarter of the points.
}
\label{fig:robustness}
\vspace{-14pt}
\end{figure}

\subsection{Robustness to Sparse Input Views}%
\label{sec:exp:e2}

To evaluate robustness to sparse input views,
we uniformly subsample each scene's saved views to $N_{\mathrm{views}} \in \{2, 4, 6, 8, 10\}$ and report the same metrics as in \Cref{tab:e1} (\Cref{tab:e2}), comparing against DA3~\citep{lin25depth}, NKSR~\citep{huang2023neural} and Surflo~\citep{guedon2026surflo} as representative baselines.
\rev{\Oursname{} leads $\mathrm{CD}_{\mathrm{all}}$ and the F-scores at every view count we test, against DA3, NKSR and Surflo}, with the largest gains in the sparsest setting: \new{the gap to NKSR in $\mathrm{CD}_{\mathrm{all}}$ is widest at $N{=}2$ on both datasets.} It also degrades less as views decrease, maintaining higher F-scores from $N{=}10$ to $N{=}2$ than visible-surface baselines. 
\revm{At $N{=}10$, its F$_{0.05}$ is 0.829 \vs 0.754 for NKSR on Replica and 0.697 \vs 0.674 on SCRREAM.}
$\mathrm{CD}_{\mathrm{vis}}$ remains slightly better for DA3, especially at very sparse views, since it reconstructs only observed surfaces while our model generates complete geometry. \Cref{fig:vis_sparse} shows that, as views are removed, the DA3 cloud becomes increasingly incomplete whereas our reconstruction remains complete.
In addition, \Cref{fig:vis_diverse} in the appendix shows that \Oursname{} can generate diverse plausible completions from the same partial observation, suggesting that it does not simply memorize a single dataset-specific shape prior.

\subsection{Ablation Study}%
\label{sec:exp:ablation}

\rev{We ablate on Replica}~\citep{straub19the-replica} to analyze the contributions of key design choices in \Oursname{}.

\heading{Importance of Partial-point Conditioning}
\new{Partial-point conditioning anchors the generative prior through two complementary paths (\Cref{sec:vpc}), ablated in \Cref{tab:ablation}(a): \emph{global anchor tokens}, encoded from the input cloud and read via cross-attention, \rev{and \emph{local spatial injection}, which injects the visible cloud into the structure latent. Removing local injection causes the largest drop, especially in visible-region fidelity ($\mathrm{CD}_{\mathrm{vis}}$), since per-voxel evidence localizes the observed geometry. Removing global anchor tokens causes a smaller drop, as the tokens supply scene-level context rather than localization.}}

{\revmcolor
\heading{Coarse-to-Fine Levels}
\Cref{tab:ablation}(b) and \Cref{fig:octree_depth} evaluate the pipeline after levels 0, 1, and 2 on five held-out real-world scenes and one Matterport3D house. Each finer level halves the mesh-cell size and recovers more geometric detail, substantially improving both the overall and high-curvature F-scores while reducing the accuracy error.
}

{\revcolor
\heading{Robustness to Imperfect Inputs}
All results above condition on raw DA3 clouds without confidence filtering. We additionally perturb the inputs without retraining (\Cref{fig:robustness}). \revm{Performance remains stable when retaining only a quarter of the input points and under Gaussian noise up to approximately the voxel scale. More severe sparsification or noise causes a gradual decline, consistent with the model's reliance on spatially accurate anchors.} Similar trends on ground-truth and DA3 points indicate that these results reflect the model's response to the perturbations rather than an interaction with DA3-specific errors.
}

\section{Conclusion}%
\label{sec:conclusion}

We presented \Oursname, a generative framework for recovering complete scene meshes from partial point clouds. Our approach combines partial-point conditioning with \revm{metric-scale coarse-to-fine generation on an octree of fixed-size windows, which keeps the mesh resolution fixed in metres regardless of scene extent and generates} complete scene geometry \revm{directly} in the original scene coordinate frame. \Oursname{} improves complete-scene reconstruction over the baselines while maintaining high fidelity to the observed geometry. We further demonstrate robustness to sparse-view inputs and strong performance on real-world scenes, highlighting its potential for applications that require complete digital representations of physical environments.

\heading{Acknowledgements}
We would like to thank Felix Wimbauer and Shenhan Qian for their insightful discussions.
This work was supported by the ERC Advanced Grant “SIMULACRON” (agreement \#884679), the GNI Project “AI4Twinning”, and the DFG project CR 250/26-1 “4DYoutube”.
Chuanxia Zheng is supported by NTU SUG-NAP and National Research Foundation, Singapore, under its NRF Fellowship Award NRF-NRFF17-2025-0009.
Andrea Vedaldi is supported by a Royal Society Faraday Discovery Fellowship FDF\textbackslash{}S2\textbackslash{}251003.

\newpage

\subsection*{AI use statement}
We used generative AI tools only to suggest local improvements to grammar and phrasing during manuscript preparation. We did not use generative AI tools for research ideation, experimental design, data analysis, interpretation of results, or the formulation of scientific claims. All AI-assisted edits were reviewed and revised by the authors, who take full responsibility for the final content of this work.

\subsection*{Ethics statement}
This work addresses complete scene-mesh reconstruction from partial point clouds,
evaluated on publicly available research datasets (InternScenes, SAGE, Replica,
SCRREAM). We are not aware of ethical concerns specific to this research, including
risks to human subjects, dataset consent issues, or foreseeable dual-use harms beyond
those already associated with general-purpose 3D reconstruction and generative
modeling research.

\subsection*{Reproducibility statement}
We describe the model architecture, training data construction, and evaluation
protocol in \Cref{sec:method,sec:experiment} of the main text, with the camera
sampler in \Cref{sec:supp:data} and the remaining implementation details, metric
definitions, and evaluated scene subsets in \Cref{sec:supp:impl}. We plan to release code and
pretrained checkpoints upon publication to support independent verification of the
reported results.

\appendix

\newpage
\section{Appendix}%
\label{sec:supp}

\subsection{Training Data Construction Details}
\label{sec:supp:data}

For each \texttt{InternScenes} room we place $1$--$9$ camera anchors by floor area, in collision-free floor locations, and render three pitch rings at $-10^\circ$, $-45^\circ$, and $+50^\circ$ from each,
with $10$ uniformly spaced yaw angles per ring, giving $30$ frames per anchor. Zenith and nadir
views and corner cameras facing the room interior cover the boundaries that centre-facing views
miss. The sampler yields approximately $40$ to roughly $420$ frames per scene depending on room size, $106$ on
average; training draws random subsets of these to form the partial point cloud.

\subsection{More Implementation Details}%
\label{sec:supp:impl}

\paragraph{Stage-1.}
The Stage-1 denoiser operates on the $16^3 \times 8$ TRELLIS.2 sparse-structure latent, with global
self-attention, rotary position embeddings and adaptive layer-norm time conditioning.
Cross-attention attends to the 768 anchor tokens of the Utonia/Q-Former encoder. Two modifications
make it visibility-aware: (i) the warm start of \Cref{eq:warmstart}, which voxelizes the input cloud,
runs the frozen TRELLIS.2 sparse-structure encoder to obtain $z_\text{vis} \in \mathbb{R}^{8 \times 16^3}$,
and replaces the Gaussian initial noise at visible voxels by $z_\text{vis}$; and (ii) a
\textsc{VisibleMaskConcat} adapter, which concatenates the binary visibility mask with the noisy latent at
each denoising step and maps the result back to 8 channels with an identity-initialized $3$D convolution.
After 12 flow-sampling steps with classifier-free guidance (scale 5, applied for $t \in [0.6, 1]$), the
latent is decoded by the unmodified TRELLIS.2 sparse-structure decoder into a $64^3$ occupancy field,
max-pooled to the $32^3$ active coordinates that are passed to Stage 2.

\paragraph{Stage-2.}
The Stage-2 denoiser is \emph{sparse}: it operates directly on the active coordinates produced by
Stage 1, one token per active cell of the $32^3$ grid, and generates the 32-channel O-Voxel latent.
Cross-attention reads anchor tokens from its own encoder of the same design. A sparse visibility
injector adds the latents of the visible voxels, with a per-voxel visibility flag, at every denoising
step, and on observed voxels the initial sample is the shape latent encoded from the visible points.
After 12 flow steps with classifier-free guidance (scale 3, applied for $t \in [0.6, 1]$), the
unmodified TRELLIS.2 decoder turns the latent into an O-Voxel field, from which the mesh is extracted
on a $512^3$ grid.

\paragraph{\revb{Metric definitions.}}
\revb{Let $\mathcal{P}$ be points sampled from the predicted mesh, $\mathcal{G}$ points sampled from the
complete ground-truth mesh, and $\mathcal{G}_{\mathrm{vis}}$ the ground-truth points returned by raycasting
the input cameras, so $\mathcal{G}_{\mathrm{vis}}$ is occlusion-aware. Let $\mathcal{P}_{\mathrm{vis}}$ be
the subset of $\mathcal{P}$ inside the union of the input view frustums; this is a geometric frustum test,
not an occlusion test, since occlusion of the prediction cannot be resolved without raycasting it. With
$d(x,\mathcal{S}) = \min_{y \in \mathcal{S}} \lVert x - y \rVert$,}
\begin{equation*}
\begin{aligned}
  a &= \tfrac{1}{|\mathcal{P}_{\mathrm{vis}}|}\!\!\sum_{x \in \mathcal{P}_{\mathrm{vis}}}\!\! d(x, \mathcal{G}),
  &\quad
  c_{\mathrm{vis}} &= \tfrac{1}{|\mathcal{G}_{\mathrm{vis}}|}\!\!\sum_{y \in \mathcal{G}_{\mathrm{vis}}}\!\! d(y, \mathcal{P}),
  &\quad
  c_{\mathrm{all}} &= \tfrac{1}{|\mathcal{G}|}\!\sum_{y \in \mathcal{G}}\! d(y, \mathcal{P}), \\[2pt]
  \mathrm{CD}_{\mathrm{vis}} &= \tfrac{1}{2}(a + c_{\mathrm{vis}}),
  &\quad
  \mathrm{CD}_{\mathrm{all}} &= \tfrac{1}{2}(a + c_{\mathrm{all}}),
  &\quad
  \mathrm{F}_{\tau} &= \tfrac{2 P_{\tau} R_{\tau}}{P_{\tau} + R_{\tau}},
\end{aligned}
\end{equation*}
\revb{with $P_{\tau} = |\{x \in \mathcal{P}_{\mathrm{vis}} : d(x,\mathcal{G}) < \tau\}| / |\mathcal{P}_{\mathrm{vis}}|$
and $R_{\tau} = |\{y \in \mathcal{G} : d(y,\mathcal{P}) < \tau\}| / |\mathcal{G}|$. Precision is therefore
restricted to the observed region while recall covers the scene, so geometry generated outside the frustums
is not charged. $\mathrm{F}_{\tau}$ is thus a completion-oriented score rather than the standard symmetric F-score,
and $\mathcal{P}_{\mathrm{vis}}$ is non-empty in all our evaluations. We sample $10^5$ points per mesh.}

\paragraph{Evaluation Datasets.}
For Replica~\citep{straub19the-replica}, we use 8 of the 18 scenes, following~\citet{ni2025decompositional}.
For SCRREAM~\citep{jung2024scrreamscanregister}, we exclude scenes whose videos focus only on a single corner of the scene.

\subsection{Additional Qualitative Results}
\label{sec:supp:more-qualitative}

\begin{figure}[t]
\centering
\includegraphics[width=\linewidth]{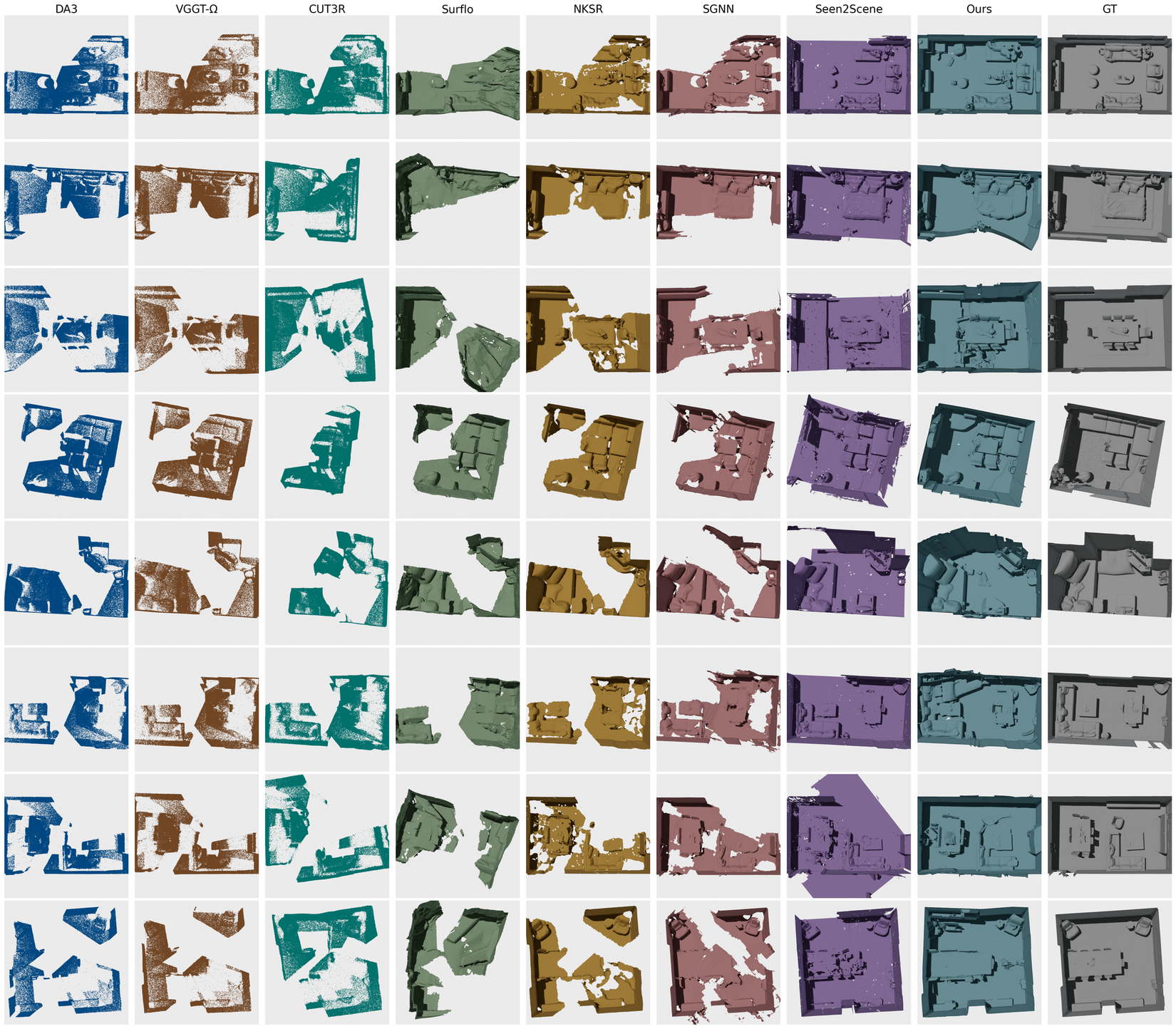}
\caption{\textbf{Additional qualitative results on Replica~\citep{straub19the-replica} at $N{=}4$.}
Top-down views of the meshes each method reconstructs from the same partial DA3 point cloud, with the ground truth in the last column.}
\label{fig:vis_replica}
\end{figure}

\begin{figure}[t]
\centering
\includegraphics[width=\linewidth]{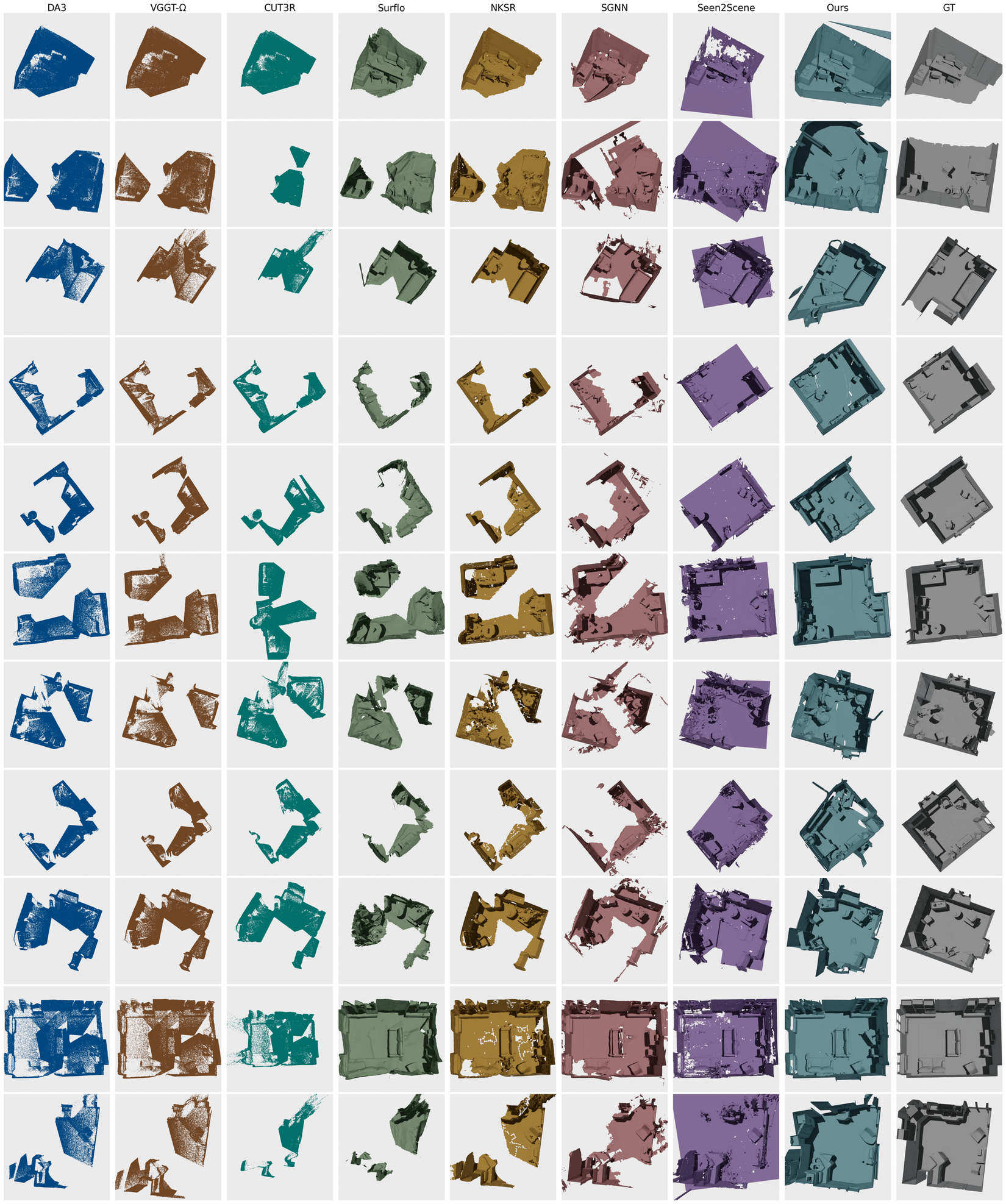}
\caption{\textbf{Additional qualitative results on SCRREAM~\citep{jung2024scrreamscanregister} at $N{=}4$.}
Top-down views of the meshes each method reconstructs from the same partial DA3 point cloud, with the ground truth in the last column.}
\label{fig:vis_scrream}
\end{figure}

\paragraph{More reconstructions on Replica and SCRREAM.}
We provide additional top-down qualitative reconstructions on Replica~\citep{straub19the-replica} and SCRREAM~\citep{jung2024scrreamscanregister} at $N_{\mathrm{views}}{=}4$ in \Cref{fig:vis_replica,fig:vis_scrream}, respectively, comparing the input partial point cloud against representative baselines and \Oursname{}.
Across both datasets, our method recovers a complete room mesh, including floors, walls, and large objects in regions where the partial input has no coverage, while baselines leave visible holes wherever the cameras did not directly observe.

\begin{figure}[t]
\centering
\includegraphics[width=\linewidth]{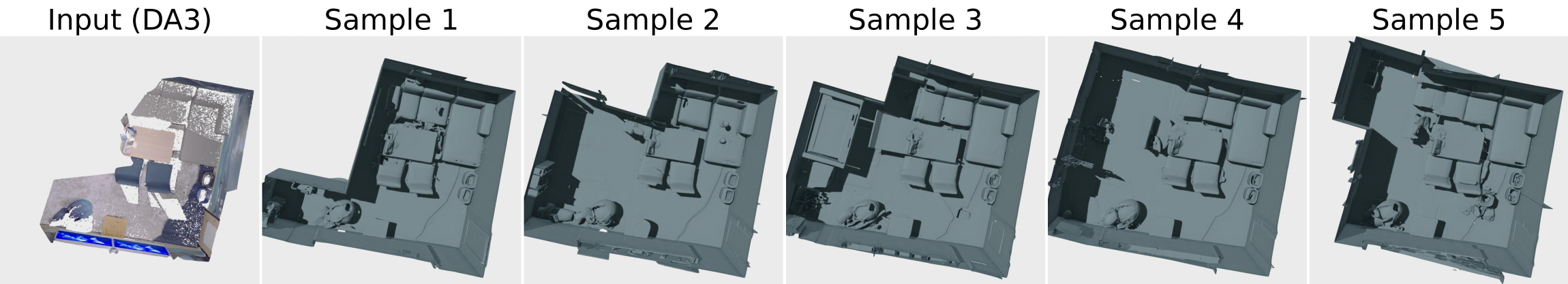}
\caption{\textbf{Generation diversity.} The input point cloud and five samples drawn from it with different seeds, rendered at level 0 with a model
trained on InternScenes and SAGE. The observed structure is kept in every sample, and the samples differ only in
the regions the input does not cover.}
\label{fig:vis_diverse}
\end{figure}

\paragraph{Generation diversity.}
\Cref{fig:vis_diverse} runs the model with different seeds on one partial point cloud. Every sample keeps the observed walls and furniture; the variation sits in the parts the input does not cover.

{\revcolor
\subsection{Generalization Across Input Sources}%
\label{sec:exp:sources}

The experiments above condition on point clouds produced by DA3~\citep{lin25depth}, a per-view point-map estimator. Because \Oursname{} conditions directly on points rather than on images or estimator-specific features, it is not architecturally tied to this source. To test this, we apply \Oursname{}, with no retraining or source-specific tuning, to partial point clouds from four acquisition paradigms: learned per-view point maps (DA3, Pi3~\citep{wang2025pi}, VGGT-$\Omega$~\citep{wang26vggt-omega}), classical structure-from-motion and multi-view stereo (COLMAP~\citep{schonberger16structure-from-motion}), monocular visual SLAM (DROID-SLAM~\citep{teed21droid-slam:}, DPVO~\citep{teed22deep}), and a simulated 32-beam LiDAR scan. All inputs are registered into the same evaluation frame and scored with the protocol used for the default DA3 setting.

\Cref{tab:sources} shows that \Oursname{} remains effective across every source we test, with F$_{0.05}$ ranging from $0.547$ to $0.761$. \revb{VGGT-$\Omega$ is the weakest of them and LiDAR the strongest, with DROID-SLAM close behind; \Oursname{} handles all of them without tuning.} In particular, the SfM/MVS, visual-SLAM, and LiDAR rows confirm that the method is not restricted to the per-view point-map setting used elsewhere in the paper.
}

\begin{table*}[t]
\centering
\footnotesize
\setlength{\tabcolsep}{4pt}
\caption{\textbf{Generalization across input point-cloud sources on Replica~\citep{straub19the-replica}.}
\Oursname{} is evaluated without retraining on partial point clouds from multiple acquisition paradigms, using the default DA3 evaluation protocol (\Cref{tab:e1}). Visual-SLAM and LiDAR use full trajectories/scans and therefore demonstrate source generality rather than budget-matched performance. $^\dagger$ LiDAR uses conservative $5\,\mathrm{cm}$ range noise.}%
\label{tab:sources}
\begin{tabular}{@{}ll cccc@{}}
\toprule
\textbf{Category} & \textbf{Input source} & $\mathrm{CD}_{\mathrm{vis}}$ $\downarrow$ & $\mathrm{CD}_{\mathrm{all}}$ $\downarrow$ & F$_{0.1}$ $\uparrow$ & F$_{0.05}$ $\uparrow$\\
\midrule
\multirow{3}{*}{Per-view point map}
 & DA3 (default)       & \revm{0.055} & \revm{0.067} & \revm{0.819} & \revm{0.684} \\
 & Pi3                 & \revm{0.048} & \revm{0.062} & \revm{0.837} & \revm{0.655} \\
 & VGGT-$\Omega$ & \revm{0.068} & \revm{0.079} & \revm{0.746} & \revm{0.547} \\
\midrule
SfM/MVS & COLMAP & \revm{0.069} & \revm{0.084} & \revm{0.770} & \revm{0.641} \\
\midrule
\multirow{2}{*}{Visual SLAM}
 & DROID-SLAM & \revm{0.044} & \revm{0.053} & \revm{0.870} & \revm{0.759} \\
 & DPVO              & \revm{0.046} & \revm{0.055} & \revm{0.872} & \revm{0.713} \\
\midrule
LiDAR & 32-beam, $\sigma{=}5\,\mathrm{cm}^\dagger$ & \revm{0.037} & \revm{0.043} & \revm{0.949} & \revm{0.761} \\
\bottomrule
\end{tabular}
\end{table*}

{\revcolor
\subsection{Failure Cases}%
\label{sec:exp:failure-cases}

\Cref{fig:robustness} perturbs the point cloud synthetically, which is not how a depth estimator
actually fails. To probe that, we degrade the \emph{images} DA3 consumes and re-run the full
pipeline on the resulting clouds. \Cref{fig:da3_failure} shows a SCRREAM scene under low light
and a real-world scene under motion blur and sensor noise, with both degradations synthetically applied to real captures. Neither scene is in our training set.

The degraded imagery costs DA3 large parts of the room, and the reconstruction inherits exactly
those gaps: \Oursname{} still closes the surface it is given and produces a coherent room, but it
completes around the missing observations rather than recovering them.
\rev{The model is trained only on clean geometry, so it treats the input as a reliable anchor and preserves systematic upstream error rather than repairing it. Training with degraded inputs is a natural remedy we leave for future work.}

{\revcolor
\begin{figure}[t]
\centering
\includegraphics[width=0.85\linewidth]{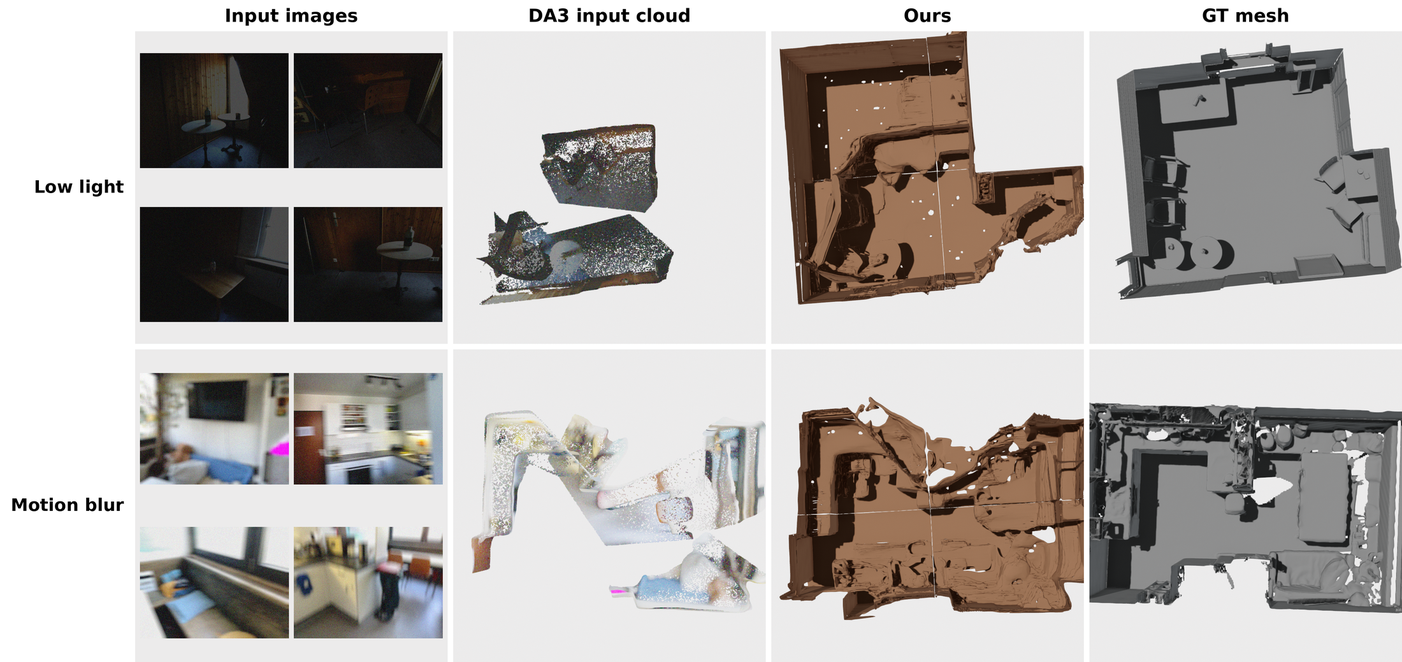}
\caption{\textbf{Failure cases under degraded imagery.} Real captures with simulated low light (top) and motion blur/noise (bottom). The degraded inputs cause DA3 to miss large scene regions, and our model completes around these gaps rather than recovering them. }%
\label{fig:da3_failure}
\end{figure}
}

\subsection{Limitations}%
\label{sec:exp:failure}
While \Oursname{} is robust to sparse input views (\Cref{sec:exp:e2}), it still has several limitations. First, the current model predicts geometry only: we use the inherited mesh decoder and do not model appearance. Extending the framework with RGB supervision and an appearance-aware decoder could enable textured mesh generation. Second, performance depends on the quality of the upstream geometry, so drift, scale errors, or systematic depth bias may propagate to the generated mesh. \Cref{sec:exp:failure-cases} shows the sharper form of this: severe upstream failures are inherited rather than corrected. Third, our training and evaluation focus on indoor scenes, and generalization to outdoor, unbounded, or substantially different scene distributions remains open.

\end{document}